\documentclass[11pt,a4paper]{article}
\usepackage[hyperref]{acl2020}
\usepackage{times}
\usepackage{latexsym}
\usepackage{amssymb}

\usepackage{microtype}
\usepackage{booktabs}
\usepackage{verbatim}
\usepackage{amsmath}
\usepackage{graphicx}
\usepackage{makecell}
\usepackage{multirow,multicol}
\usepackage{colortbl}
\newcommand{\cg}[1]{\cellcolor{lightgray}#1}
\usepackage[disable]{todonotes}

\newcommand\blfootnote[1]{%
  \begingroup
  \renewcommand\thefootnote{}\footnote{#1}%
  \addtocounter{footnote}{-1}%
  \endgroup
}
\usepackage{xcolor}

\title{Multilingual Speech Translation \\ with Efficient Finetuning of Pretrained Models}

\author{Xian Li*, Changhan Wang*, Yun Tang, Chau Tran, Yuqing Tang, Juan Pino, \\ \textbf{Alexei Baevski, Alexis Conneau, Michael Auli} \\
 Facebook AI\\
\texttt{\{xianl,changhan,yuntang,chau,yuqtang,juancarabina,}\\\texttt{abaevski,aconneau,michaelauli\}@fb.com}         
}

\begin{document}
\maketitle
\
\blfootnote{\textbf{*} Equal contribution.}
\begin{abstract}

We present a simple yet effective approach to build multilingual speech-to-text (ST) translation by efficient transfer learning from pretrained speech encoder and text decoder. Our key finding is that a minimalistic LNA (\textbf{L}ayer\textbf{N}orm and \textbf{A}ttention) finetuning can achieve zero-shot crosslingual and cross-modality transfer ability by only finetuning less than 10\% of the pretrained parameters. This enables effectively leveraging large pretrained models with low training cost. Using wav2vec 2.0 for acoustic modeling, and mBART for multilingual text generation, our approach advanced the new state-of-the-art for 34 translation directions (and surpassing cascaded ST for 23 of them) on large-scale multilingual ST benchmark CoVoST 2 ~\cite{wang2020covost} ($+6.4$ BLEU on average across 15 En-X directions and $+5.1$ BLEU on average across 19 X-En directions). Our approach demonstrates strong zero-shot performance in a many-to-many multilingual model ($+5.7$ BLEU on average across 18 non-English directions), making it an appealing approach for attaining high-quality speech translation with improved parameter and data efficiency.

\end{abstract}

\section{Introduction}
\label{sec:intro}

Recent advances in pretraining over unlabeled data and then finetuning on labeled data leads to significant performance improvement in text understanding and generation tasks \cite{bert, Liu2020MultilingualDP, conneau2019unsupervised,  Radford2018ImprovingLU}. Lately, such text pretraining and finetuning paradigms have been extended to other modalities: audio \cite{schneider2019wav2vec, Baevski2020wav2vecV2}, images \cite{su2019vl,lu2019vilbert}, and video \cite{sun2019videobert}. At the same time, pretraining and finetuning techniques have improved multi-tasking applications significantly, such as multilingual translation, cross-lingual representations, question-answering and so on \cite{raffel2020exploring, yang2019xlnet, Tang2020MultilingualTW}. In this paper, we advance the one-model-for-all paradigm further by adapting audio and multilingual text pretraining and finetuning to improve multilingual speech-to-text translation. 

\begin{figure}[t]
    \centering
    \includegraphics[width=1.0\linewidth]{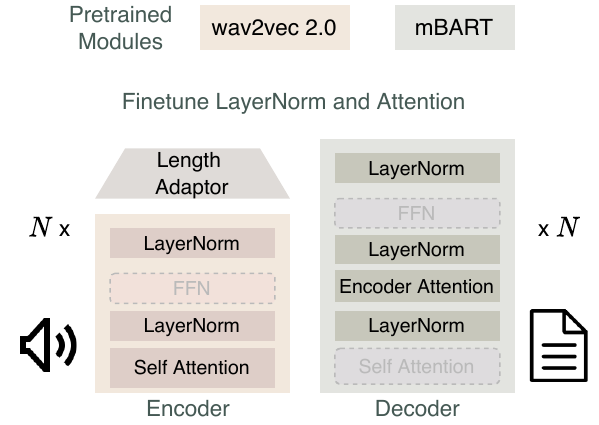}
    \caption{An overview of the proposed speech-to-text translation via transfer learning from efficient finetuning of single-modality pretrained models. The proposed LNA finetuning is applied to each layer.}
    \label{fig:overview}
\end{figure}

Our contributions are as follows:
\begin{itemize}
    \item We propose a simple and effective approach to combine pretrained single-modality modules to perform speech-to-text translation. With minimal architecture change, we added a cross-modal adaptor to bridge the length discrepancy between audio encoder output and text decoder input. Our approach can also perform multi-task finetuning with both speech-to-text translation and text-to-text translation tasks where we found joint training with the latter brings further gains.
    \item We present an efficient transfer learning strategy by only finetuning the  \textbf{L}ayer\textbf{N}orm and \textbf{A}ttention (LNA) parameters of pretrained models. This approach is not only parameter- and data-efficient but also effective for zero-shot crosslingual transfer to unseen languages (train on A $\rightarrow$ B, test on A $\rightarrow$ C and C $\rightarrow$ B).
    \item Our approach is also effective for zero-shot multilingual translation (train on A $\rightarrow$ B and B $\rightarrow$ C, test on A $\rightarrow$ C), which provides a cost-effective approach for many-to-many speech-to-text translation without dependency for parallel data for every direction. 
    
    \item Using a pretrained audio encoder (wav2vec \cite{Baevski2020wav2vecV2}) and multilingual text decoder (mBART~\cite{Liu2020MultilingualDP}), this approach sets new state-of-the-art (SOTA) on two large-scale speech translation benchmarks. On CoVoST 2 \cite{wang2020covost}, we pushed the SOTA for end-to-end approach for 19 X-En directions($+5.1$ BLEU on average) and 15 En-X directions ($+6.4$ BLEU on average) by finetuning only $10 \sim 50$\% of parameters. Similarly on Europarl \cite{jairsan2020a}, our zero-shot multilingual many-to-many model is not only data efficient, but also bring $+5.7$ BLEU (on average) when translating 18 non-English directions compared to a many-to-many model training on 1.6$\times$ training data (all pair-wise directions).

\end{itemize}

We describe our approach in Section \ref{sec:modules}, namely pretrained models, length adaptor, LNA finetuning and joint speech-text finetuning. Experiments setup and results are elaborated in Section \ref{sec:exp} and Section \ref{sec:results}. Section \ref{sec:ablation} provides ablation studies of the proposed finetuning strategy.


\section{Methods}
\label{sec:modules}

\subsection{Pretrained Modules}
Our model leverages a pretrained encoder from wav2vec 2.0~\cite{Baevski2020wav2vecV2} for acoustic modeling and a pretrained decoder from multilingual BART (mBART)~\cite{Liu2020MultilingualDP} for language modeling.

\noindent {\bf wav2vec 2.0} is a simple and powerful framework to learn high quality speech representation from unlabelled audio data. It mainly consists of two components: feature encoder and context encoder. The feature encoder, which is built from temporal convolution layers, takes raw audio signal $O$ as input and generates latent speech representation $Z=[z_1,\cdot\cdot\cdot,z_T]$. They are fed to the transformer based context encoder to generate context representations $C=[c_1,\cdot\cdot\cdot,c_T]$ with sequence level information. During pre-training, the model is optimized with a contrastive task to distinguish true latent from distractors. The input to the context encoder is with span masked. The latent speech representation $Z$ is discretized to $Q=[q_1, \cdot\cdot\cdot, q_T]$ and used as targets for the frames in the masked span. 

\noindent {\bf mBART} is a sequence-to-sequence generative pretraining scheme, specifically a denoising autoencoder (DAE) to predict the original text $x$ given $g(x)$ where $g$ is a noising function that corrupts text such as random span masking and order permutation \cite{Liu2020MultilingualDP}. The model is trained with monolingual data of $N$ languages: 
$\mathcal{D}=\{\mathcal{D}_1, ..., \mathcal{D}_N \}$ where each $\mathcal{D}_i$ is a collection of documents in language $i$. The pretraining objective optimizes $\mathcal{L}_\theta$: 
\begin{equation}
    \mathcal{L}_\theta = 
    \sum_{\mathcal{D}_i\in \mathcal{D}}
    \sum_{x\in \mathcal{D}_i}\log P(x | g(x); \theta)~,
    \label{eq:learning}
\end{equation}
where $x$ is an instance in language $i$ and the distribution $P$ is parameterized by the sequence-to-sequence model.


\subsection{Length Adaptor}
\label{sec:adaptor}
We add a lightweight adaptor module in between encoder and decoder to better align the two modules pretrained with different modalities. The adaptor module performs projection and downsampling to alleviate length inconsistency between the audio and text sequences. Specifically, we use a stack of $n$ 1-dimensional convolutional layers with stride 2 to shrink the speech sequence (encoder output) by a factor of $2^n$.



\subsection{LNA Finetuning}
Compared to the baseline approach of finetuning all parameters in pretrained models, we propose parameter efficient finetuning strategy (\textbf{LNA}) of only finetuning the layer normalization (\textbf{L}ayer\textbf{N}orm) and multi-head attention (MH\textbf{A}) parameters. LNA is motivated to bridge the discrepancy between pretraining and downstream (ST) task, which we hypothesize are accounted by the following parameters:

\noindent \textbf{LayerNorm} parameters from pretrained models were trained based on the statistics of the data used in pretraining and thus need to be adapted to downstream tasks during finetuning. 

\noindent \textbf{Attention} Encoder attention (EA, attention to encoder outputs) parameters from pretrained MT decoder were trained on the text-to-text MT task, so we hypothesize that they are crucial to be adapted to the speech encoder output. Combined with LayerNorm parameter is the proposed LNA-Minimalist finetuning. In addition, we also investigate the role of self attention (SA) parameters in facilitating crosslingual transfer ability.






\subsection{Joint Speech-text Finetuning}
Multi-task learning has shown to be an effective approach to improve the performance of the speech translation task using other related tasks, such as MT and ASR~\cite{Weiss2017SequencetoSequenceMC,Anastasopoulos2018TiedML,Bahar2019ACS,Tang2020AGM}. We jointly train MT and ST tasks in the finetuning with pre-trained models. The speech transcripts is used as input for the MT task and the corresponding speech data is used as input for the ST task. As a result, we can leverage abundant parallel text data to further improve the performance.



\section{Experimental Setup}
\label{sec:exp}
\subsection{Datasets}
We evaluate our proposed models on two large-scale multilingual speech translation benchmarks. 
\textbf{CoVoST 2} ~\cite{wang2020covost} is a multilingual speech-to-text translation corpus with English into 15 languages (En-X) and 21 languages into English (X-En). It provides a comprehensive test bed for low-resource scenarios, with 4 X-En directions between 10 hours and 20 hours training data, and 11 X-En directions less than 4 hours training data.  

\noindent \textbf{Europarl ST} \cite{jairsan2020a} has both English-centric as well as non-English directions, which allow us to evaluate the proposed method's effectiveness of multilingual translation between any pair especially zero-shot performance. We experiment on all 6 languages (de, en, es, fr, it, pt). We compare to a multilingual baseline trained with all pair-wise parallel data.

\subsection{Training}

We evaluate the following instantiation of the proposed method which is referred to as XMEF (Cross-Modal Efficient Finetuning):

\noindent \textbf{Encoder.} We initialize the encoder using the opensourced wav2vec 2.0 large architecture pretrained on unlabelled English-only (w2v-En) audio from LibriVox \cite{Baevski2020wav2vecV2}. Encoder output is followed by 3 1-D convolution layers with stride 2 to achieve 8x down-sampling of audio encoder outputs. 

\noindent \textbf{Decoder.} We use the same pretrained mBART models from  \cite{Tang2020MultilingualTW}. It was first pretrained monolingual data of 50 languages (from CommonCrawl) and then continued to train with bitexts of $49$ languages. We use MBART-ML50N1 ($49$ languages to English) for X-En ST directions and MBART-ML501N (English to $49$ languages )for translating En-X ST directions.  

\noindent \textbf{LNA Finetuning.} We study the parameter efficiency and crosslingual transfer ability of  LNA finetuning in bilingual setting without the additional effect from multilingual training. Drawing learnings on that, we then evaluate applying LNA finetuning to encoder only (LNA-E), decoder only (LNA-D), and both (LNA-E,D) respectively. For multilingual finetuning on CoVoST 2, we use all X-En training data (except zero-shot crosslingual transfer experiments) for evaluating X-En performance, and En-X data from all directions for evaluating En-X performance. For evaluating multilingual zero-shot performance on Europarl, we only use X-En and En-X for finetuning and evaluate on all (X-X) pairs.

\begin{figure}[t]
    \centering
    \includegraphics[width=1.1\linewidth]{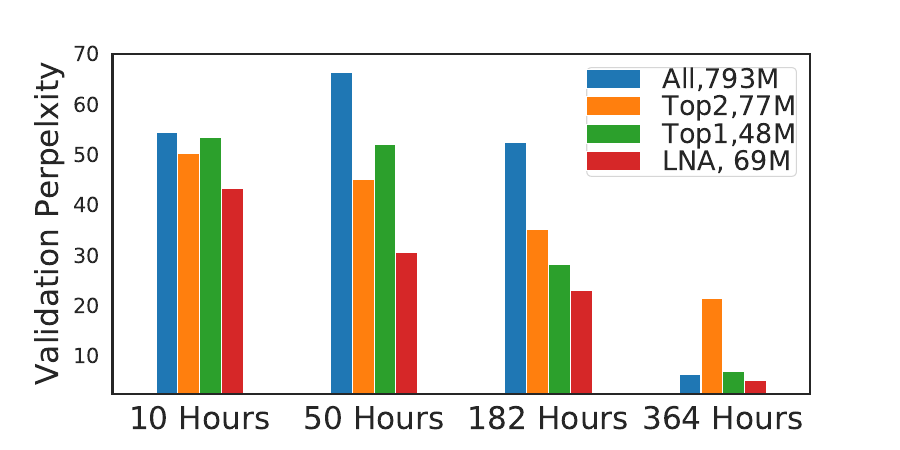}
    \caption{Comparison of LNA finetuning with alternative finetuning strategies: finetuning all parameters (All), finetuning top-$k$ layers (Top1, Top2). We evaluate generalization (perplexity on dev set) performance with different amount of training data. LNA achieves the best generalization with substantially less parameters. Experiments are done using CoVoST En-De. } 
    \label{fig:lna}
\end{figure}

\begin{figure}[t]
    \centering
    \includegraphics[width=0.9\linewidth]{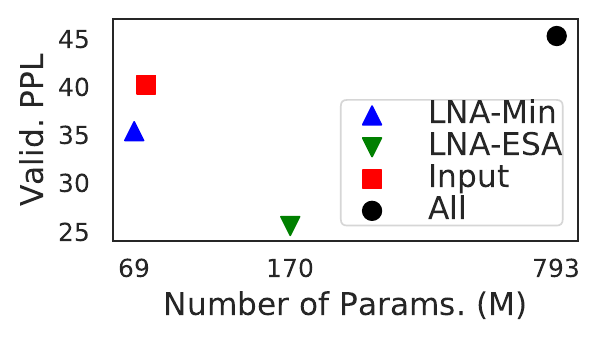}
    \caption{Comparison of LNA finetuning with alternatives: finetuning all parameters (All) and finetuning feature extractor (Input), on adapting wav2vec English encoder to translate non-English speech input. Experiments are done using CoVoST De-En.} 
    \label{fig:lna-esa}
\end{figure}
\noindent \textbf{Joint Training.}  Two encoders are initialized with pre-trained mBART encoder and wav2vec encoder mentioned above, and are used for text and speech input respectively. The last 12 transformer layers in the wav2vec encoder is replaced with 12 mBART encoder layers. Parameters in those 12 layers are shared between two encoders during joint training.  The decoder is also shared between two tasks and is initialized with the pre-trained mBART decoder model. We also experimented with adding additional bitext used in ML50~\cite{Tang2020MultilingualTW} as training data for the MT task.  Only language pairs exist in the CoVoST 2 data set are chosen and they cover all language pairs in the CoVoST 2 data set except English to and from ``Ca" and ``Cy".  We fine-tune all parameters in this experiments since the big mismatch of the pre-trained model (mBART encoder as part of the speech encoder) and more available training data.

\subsection{Baselines}

\noindent {\bf From scratch:} The first baseline trains a sequence-to-sequence model with Transformer architecture without any pretraining.

\noindent {\bf ASRPT+Multi:} Pretraining encoder on ASR task was shown to be an effective method to improve speech translation and accelerates convergence ~\cite{bansal-etal-2019-pre}. We compare our results to a strong baseline provided by \cite{wang2020covost}, consisting of a multilingual Transformer model trained on  CoVoST 2 with multilingual ASR pretraining (ST).  

\noindent {\bf XMEF-BL:} Multilingual model for En-X (one-to-many, when the model needs to translate to multiple target languages) usually faces more challenges from interference which has been observed in general multi-task learning \cite{lin2019pareto} as well as multilingual machine translation for text  and renders them not always outperforming bilingual counterparts \cite{arivazhagan2019massively}. Therefore, we compare to applying our method (XMEF, LNA) to bilingual finetuning i.e. finetuned on parallel data from a single language pair.

\begin{table*}[h]
\centering
\begin{tabular}{r|ccc|ccccc|c}
\toprule
\multicolumn{4}{c|}{ \multirow{2}{*}{} }   & \multicolumn{5}{c|}{\bf Train } & \multicolumn{1}{c}{\bf  Zero-shot}  \\
\midrule
  &  Enc & Dec & Params. & Fr &De & Es & Ca & It & Pt \\ 
\midrule
LNA-E,D &   LN+SA & LN+EA & 170.7M &  \bf 32.4 & \bf 24.9  & \bf 31.6 & \bf 28.6 &  \bf 24.0 &  \bf 8.2 \\ 
LNA-D &  All &  LN+EA & 384.8M &  31.6 &  23.7 &  31.0 &  27.8 &  23.2 &   7.6 \\ 
Finetune All &  All &  All & 793.0M & 27.1  & 17.7 &  27.8 & 21.7  & 18.9  & 5.1 \\ 
 
 \midrule
\multicolumn{4}{r|}{ASRPT+Multi} & 23.1  & 15.3 & 21.2 & 19.9 & 14.9 & 4.4 \\ 
\multicolumn{4}{r|}{Supervised (Multi) SOTA ~\cite{wang2020covost}} & 26.5  & 17.6 & 27.0 & 23.1 & 18.5 & 6.3 \\ 
\bottomrule
\end{tabular}
\caption{Performance on zero-shot transfer on the source-side (speech). Each model is finetuned on 5 directions from \{\textit {Fr, De, Es, Ca, It}\}  $\rightarrow$ \textit{En}, and evaluated on \textbf{unsupervised translation} of a new language (\textit{Pt}). We report BLEU scores on test set, and compare to the zero-shot transfer performance of a supervised multilingual  baselines (ASRPT+Multi), as well as previous state-of-the-art which is also the supervised and multilingually trained.} 
\label{tab:m2o-zero}
\end{table*}

\begin{table*}[h]
\centering
\begin{tabular}{r|ccc|cccc|c}
\toprule
\multicolumn{4}{c|}{ } & \multicolumn{4}{c|}{\bf Train   } & \bf Zero-shot  \\
\midrule
 &  Enc & Dec & Params. & De & Fa & Tr & Zh  & Ja   \\
\midrule
LNA-E,D & LN & LN+EA & 69.4M & 22.1 &  17.7  &  13.4 & 29.2 &  22.9 \\
LNA-E,D & LN+SA & LN+EA & 170.7M &  23.8 &  19.2  &  14.2 &  30.6 &  29.2    \\
LNA-D & All & LN+EA & 384.8M & \bf 24.9 & \bf 19.8 &  15.2 &\bf 32.7  &  \bf 30.6   \\
LNA-E & LN+SA & All & 477.6M & 22.0 & 18.1 & 14.2 & 29.5  &  0.8   \\
Finetune All & All & All & 793.0M & 24.1 & 19.6 & \bf 15.6 & 32.4  &  0.4   \\
\midrule
 \multicolumn{4}{r|}{ASRPT+Multi} & 9.5  & 10.9 & 6.8 & 23.5 & 0.0 \\ 
 \multicolumn{4}{r|}{Supervised (Multi)  SOTA ~\cite{wang2020covost}}  & 17.3  & 14.5 & 10.7 & 28.2 & 31.9   \\

\bottomrule
\end{tabular}
\caption{Performance on zero-shot transfer on the target-side (text). Each model is finetuned on 4 directions  \textit{En} $\rightarrow$ \{\textit {De, Fa, Tr, Zh}\}, and evaluated on \textbf{unsupervised translation} to a new language (\textit{Ja}). We report BLEU scores on test set, and compare to the zero-shot transfer performance of a supervised multilingual  baselines (ASRPT+Multi), as well as previous state-of-the-art which is also the supervised and multilingually trained. }
\label{tab:o2m-zero}
\end{table*}


\begin{table*}[h]
\centering
\begin{tabular}{r|llll|llllll}
\toprule
 \multicolumn{1}{c|}{} &  \multicolumn{4}{c|}{\bf High Resource } &\multicolumn{6}{c}{\bf Low Resource }  \\
 \midrule
\multicolumn{1}{r|}{\bf $\rightarrow$ En} & {Fr} & {De}  &{  Es} &{  Ca} &{  It} & { Ru}  &{  Pt} &{ Nl} &{ Sl} &{ Sv}   \\
\multicolumn{1}{r|}{Train Hours}  & 264 & 184 & 113 & 136 & 44 & 18 & 10 & 7 & 2 & 2   \\
\midrule
\multicolumn{1}{r|}{Scratch-BL} & 24.3  & 8.4& 12.0 & 14.4 & 0.2& 1.2 & 0.5 & 0.3 & 0.3 & 0.2  \\
\multicolumn{1}{r|}{+ ASR PT} & 26.3 & 17.1 &  23.0 & 18.8 & 11.3 & 14.8 & 6.1 & 3.0 & 3.0 & 2.7  \\
\multicolumn{1}{r|}{+ Multi.} & 26.5 & 17.5 & 27.0 & 23.1 & 18.5 & 4.7 & 6.3 & 5.0 & 0.7 & 0.5 \\
\multicolumn{1}{r|}{+mBART} & 28.1 & 19.7 & 28.1 & 24.0 & 19.9 & 2.7 & 6.2 & 8.1 & 0.5 & 1.4  \\
\midrule
LNA-E,D (170.7M) & {\bf 33.8*} & {\bf 26.7* } & {\bf 34.0*} & {\bf 29.5*} & {\bf 26.1*}& {\bf 21.1} & {\bf 19.2}& {\bf 14.1*} & \bf 4.6 & \bf \underline{5.9}  \\
LNA-D (384.8M) & {\bf \underline{35.0*}} &{\bf 28.2*} & {\bf \underline{35.2*}} & {\bf \underline{31.1*}} & {\bf 27.6*}  & {\bf 22.8}& {\bf \underline{24.1*}}& {\bf 14.2*}& \bf 5.0 & \bf 5.0 \\
Finetune All (793.0M) & {\bf33.0*} & {\bf24.5*} & {\bf 33.6*} & {\bf 28.0*} &{\bf 25.2*}& \bf 20.2 & {\bf 19.5} &{\bf 9.4} & \bf 4.6 & \bf 4.8  \\
Joint Training (1.05B)  & {\bf33.5*} & {\bf28.6*} & {\bf 33.5*} & {\bf 30.6*} &{\bf 26.6*} & \bf 17.6 & {\bf 12.0} &{\bf \underline{15.0*}} & \bf 3.9  &  2.6 \\
 + Extra MT Data & {\bf 34.4*} & {\bf\underline{29.6*}} & {\bf 34.4*} & {\bf 30.6*} &{\bf \underline{27.7*}}& \bf \underline{27.7*}   & {\bf 14.6} &{\bf 14.5*} & \bf \underline{5.2} &  \bf 3.4 \\
\midrule
\multicolumn{1}{r|}{Prev. E2E SOTA} & 27.0 & 18.9 & 28.0 & 24.0 & 11.3& 14.8 & 6.1 & 8.4 & 3.0 & 2.7 \\
\multicolumn{1}{r|}{Cascade SOTA} & 29.1 & 23.2 & 31.1 & 27.2 & 22.9 & 25.0 & 22.7 & 10.4 & 7.0 & 11.9\\
\bottomrule
\end{tabular}
\caption{Performance of \textit{X} $\rightarrow$ \textit{En} multilingual model on Romance, Germantic and Slavic language families. We report BLEU scores on test set. For each XMEF method, we report the number of parameters trained in brackets. Previous E2E SOTA is the best-performing end-to-end multilingual (with ASR pretraining) model from ~\cite{wang2020covost}. Results in \textbf{bold} are where the proposed approach improves previous E2E SOTA, and sets new SOTA as is \underline{underlined}. * means our new E2E SOTA also beats the previous cascade SOTA. 
}
\label{tab:m2o-romance}

\end{table*}

\begin{table*}[h]
\centering
\scalebox{0.93}{
\begin{tabular}{r|lllllllllll|r}
\toprule
{\bf $\rightarrow$ En}   & { Fa} & {Zh} &{Tr} &{Et} &{Mn}  & {Ar} &{Lv}  & {Cy} &{Ta}   &{Ja}   &{Id}  &{Avg.}    \\
Train Hours & 49 & 10  & 4 & 3 & 3  & 2 & 2 & 2 & 2  & 1 & 1  \\
 ASR (WER)  & 62.4& 45.0 & 51.2 & 65.7 & 65.2  & 63.3 &  51.8 & 72.8 & 80.8 & 77.1 & 63.2   \\
\midrule
Baseline  & 1.9& 1.4 & 0.7  & 0.1 & 0.1 & 0.3 &  0.1 & 0.3 &  0.3 & 0.3 & 0.4  \\
+ ASR PT   & 3.7& 5.8 & 3.6  & 0.1 & 0.2 & 4.3 & 2.5 & 2.7&  0.3 & 1.5 & 2.5  \\
+ Multi. & 2.4 & 5.9 & 2.3  & 0.6 & 0.1 & 0.4 &  0.6 & 1.9 & 0.1 & 0.1 & 0.3 & 7.0  \\

+ mBART & 3.3& 5.4 & 2.4  & 0.7 & 0.2 & 0.5 &  0.6 & 1.4 & 0.1 & 0.2 & 0.2 & 7.3  \\
\midrule
LNA-E,D (170.7M) & \bf 4.0 & \bf 6.2 & \bf \underline{5.5} & \bf 1.3 & \bf \underline{1.0}  & 3.7  & \bf 4.6  & 2.8 & \bf 0.7 & \bf 1.7 & \bf 2.9 & 12.5    \\
LNA-D (384.8M) &  3.6 & \bf 6.0 & \bf 4.8   & \bf \underline{1.5}  & \bf 0.9  & 2.8  & \bf \underline{4.9}   & 2.3 & \bf \underline{0.8} & \bf 1.7 & \bf \underline{3.7} & 12.6   \\
Finetune All (793.0M) & 3.7 & \bf \underline{6.5} &  \bf 4.0 &  \bf 1.4 &  \bf 1.0  & 3.3  &  \bf 4.9 & 2.1  & \bf 0.5 & \bf \underline{2.1} & \bf 3.4 & 11.2  \\
Joint Training (1.05B)  & \bf \underline{6.1*} &  5.4 &  3.3 &  0.7 &  0.2  & 0.8 &  \bf 2.7  & 1.0  &  0.1 & 0.3 & 0.5 & 10.7  \\
 + Extra MT Data  & \bf 5.0 & \bf 6.2 &  \bf 4.0 &  0.8 &  0.3  & 1.0  &  \bf 3.6 & 1.1  & 0.2 & 0.5 & 0.5 & 11.7  \\
\midrule
Prev. SOTA & 3.7 & 5.9 & 3.7& 0.9 & 0.2  & 4.3 & 2.5 & 3.3 & 0.3 & 1.5 & 2.5  \\
Cascade & 5.8 & 11.4 & 9.3 & 3.8 & 1.0  & 12.3  & 7.2  & 7.4 & 0.4 & 3.8 & 11.8  \\ 
\bottomrule

\end{tabular}
}
\caption{Performance of \textit{X} $\rightarrow$ \textit{En} multilingual model on distant (from English) and extremely low resource languages. We report BLEU scores on test set. For each XMEF method, we report the number of parameters trained in brackets. Results in \textbf{bold} are where the proposed approach improves previous E2E SOTA, and sets new SOTA as is \underline{underlined}. * means our new E2E SOTA also beats the previous cascade SOTA. For multilingual models, we also report the average (Avg.) BLEU scores across \textit{all 21} directions. 
}
\label{tab:m2o-distant}
\end{table*}

\noindent {\bf Previous SOTAs:} We compare to the best end-to-end (E2E) model from previous literature \cite{wang2020covost,jairsan2020a} on each translation direction, which is usually the best-performing multilingual model trained with parallel data from all directions (both X-En and En-X) and also pretrained with ASR. Even though the focus of the proposed method is E2E model, we also compare to the best performing cascade approach (Cascade SOTA) which is composed of Transformer-large encoder from ASR pretraining and a multilingual MT model trained on all X-En and En-X data.

\begin{table*}[h]
\centering
\begin{tabular}{r|llllllllllll}
\toprule

\multicolumn{1}{r|}{\bf En $\rightarrow$} & {Ar} & { Ca} & { Cy} & { De} &{Et} &{ Fa} &{ Id} &{ Ja} \\
\midrule
\multicolumn{1}{r|}{Scratch-BL} & 8.7 & 20.2 & 22.2 & 13.6 & 11.1 & 11.5 & 18.9 & 26.9  \\
\multicolumn{1}{r|}{+ ASR PT} & 12.1 & 21.8 & 23.9 & 16.5 & 13.4 & 13.5 & 20.8 & 29.6 \\
\multicolumn{1}{r|}{+ Multi.} & 13.0 & 22.3 & 23.7 & 17.3 & 13.9 & 14.5 & 20.3 & 31.9 \\
\midrule
LNA-E,D-BL (69.4M) & 12.0 & 18.8 & 12.9 & \bf 20.3* & 15.0 & \bf 15.9* & \bf 24.4* & 31.4   \\
LNA-E,D  (69.4M) & {\bf 15.3*} & 20.3 & 13.2 & {\bf 23.2*} & {\bf 18.6*} & {\bf 19.6* } & {\bf 26.5* } & {\bf 36.9* } \\
LNA-E,D (170.7M) & {\bf 17.4*} & 22.2 & 14.8 & {\bf 25.3*} & {\bf 21.0*} & {\bf 20.1* } & {\bf 27.6* } & {\bf 38.4* } \\
LNA-E (477.6M) & {\bf 17.2*} & {\bf29.5*} & \bf 30.3* & {\bf 25.2*} & {\bf 20.7*} & {\bf 19.8* } & {\bf 28.5* } & {\bf 37.8* } \\
Finetune All (793.0M) & {\bf 17.7*} & \bf 30.1* & \bf 30.0* & {\bf 25.2*} & {\bf 21.1*} & {\bf 20.3* } & {\bf 28.9* } & {\bf 38.1* } \\
Joint Training (1.05B) & {\bf \underline{18.0*}} & \bf \underline{30.9*} & \bf \underline{30.6*} & {\bf \underline{25.8*}} & {\bf \underline{22.1*}} & {\bf  \underline{21.5*}} & {\bf \underline{29.9*}} & {\bf \underline{39.3*}} \\
\midrule
\multicolumn{1}{r|}{Prev. E2E SOTA} & 13.9 & 23.6 & 25.1 & 18.4 & 15.1 &15.5 & 22.0 & 33.0 \\
\multicolumn{1}{r|}{Cascade SOTA} & 14.3 & 25.0 & 25.6 & 19.4 & 15.4 & 14.1 & 23.1 & 33.8\\
\midrule
\midrule
\multicolumn{1}{r|}{\bf En $\rightarrow$}& {Lv}  &{Mn} &{Sl} &{Sv} &{Ta}  & {Tr} &{Zh} &{Avg.}  \\
\midrule
\multicolumn{1}{r|}{Scratch-BL} & 11.5 &  6.6 & 11.5 & 20.1 & 9.9 & 8.9 & 20.6 \\
\multicolumn{1}{r|}{+ ASR PT} & 13.1 & 9.2 & 16.1 & 22.3 & 11.2 & 10.2 & 25.7  \\
\multicolumn{1}{r|}{+ Multi.} & 14.1 & 10.2 & 17.1 & 22.3 & 11.7 & 10.7 & 28.2 & 18.1 \\
\midrule
LNA-E,D-BL (69.4M)  & 14.3 & 6.9 & 17.9 & \bf 26.1* & 12.6 & 10.8 & 21.8  \\
LNA-E,D (69.4M) & {\bf 17.9* } & {\bf 12.0* } & {\bf 21.1* } & {\bf 27.5* } & {\bf 14.6*} & {\bf 14.1* }& \bf 32.1* & 20.9 \\
LNA-E,D (170.7M) & {\bf 20.1* } & {\bf 13.3* } & {\bf 23.0* } & {\bf 29.6* } & {\bf 16.4*} & {\bf 15.5* }& \bf 33.0*  & 22.5 \\
LNA-E (477.6M) & {\bf 20.2* } & {\bf 14.1* } & {\bf 23.5* } & {\bf 30.0* } & {\bf 16.8*} & {\bf 16.2* }& \bf 32.8*  & 24.2  \\
Finetune All (793.0M) & {\bf 20.8* } & {\bf 14.1* } & {\bf 23.6* } & {\bf \underline{30.4*} } & {\bf 17.1*} & {\bf 16.3* }& \bf \underline{33.7*} &  24.5  \\
Joint Training (1.05B)  & {\bf \underline{21.5*}} & {\bf \underline{14.8*}} & {\bf \underline{25.1*}} & {\bf 29.6*} & {\bf \underline{17.8*}} & {\bf \underline{17.0}}& \bf 33.3 & 25.1  \\
\midrule
\multicolumn{1}{r|}{Prev. E2E SOTA} & 15.2 & 11.0 & 18.3 & 24.1 & 12.8 &11.7 & 31.3  \\
\multicolumn{1}{r|}{Cascade SOTA} & 15.6 & 11.7 & 18.9 & 24.8 & 13.7 & 11.7 & 26.9 \\

\bottomrule
\end{tabular}
\caption{Performance on \textit{En} $\rightarrow$ \textit{X} multilingual ST. We report BLEU scores on test set. For each XMEF method, we report the number of parameters trained in brackets. `BL' refers to using the same XMEF and LNA-E,D finetuning but only on bilingual corpus. Results in \textbf{bold} are where the proposed approach improves previous E2E SOTA, and sets new SOTA as is \underline{underlined}. * means our new E2E SOTA also beats the previous cascade SOTA. For multilingual models, we also report the average (Avg.) BLEU scores across \textit{all 15} directions.
}
\label{tab:o2m-seen}

\end{table*}

\section{Results}
\label{sec:results}
\subsection{Parameter Efficiency}
\label{subsec:fexl}

First, we evaluate the transfer learning performance of finetuning the entire pretrained model as well as the proposed efficient finetuning (LNA). To separate the additional crosslingual transfer learning from multilingual finetuning, we evalute on bilingual ST (En-De and De-En in CoVoST) task. We first evaluate LNA-Minimalist (69M params), and compare to finetuning all parameters, and only top layers which were found effective in transfer learning in NLP tasks with pretrained BERT \cite{wu2019beto,kovaleva2019revealing}.  Figure \ref{fig:lna} show that in both low data and high data regimes, our proposed LNA-Minimalist both generalizes better (lower perplexity on dev set) and substantially improves training efficiency (only 10\% of parameters to train leading to lower memory cost and faster trains).

\subsection{Crosslingual Transfer}
To assess transfer ability from wav2vec encoder pretrained on English to other (speech) input languages, we evaluate on CoVoST 2 De-En ST task. We investigated the role of finetuning encoder self-attention (LNA-ESA) in facilitating crosslingual transfer. We compare to baselines of finetuning the entire encoder (All), and finetuning feature extractor which are commonly used in adaptation in ASR \cite{Rivire2020UnsupervisedPT}. Results are summarized in Figure \ref{fig:lna-esa}. LNA still demonstrates improved generalization than alternative finetuning approaches, with finetuning encoder self attention (LNA-ESA) being crucial for adapting pretrained English encoder to other languages. 


\subsection{Multilingual Finetuning}
Next, we evaluate crosslingual transfer with multilingual finetuning. 

\noindent \textbf{Source-side (speech) transfer.} We evaluate whether the proposed approach enables positive crosslingual transfer to translate speech \textit{from} unseen languages in Table \ref{tab:m2o-zero}. We finetune on labelled data for 5 to-English language pairs, and evaluate the finetuned model's zero-shot performance when translating speech input from unseen languages (Pt). 

First, we found that comparing to finetuning more parameters (LNA-D, and All), LNA finetuning (LNA-E,D) not only trains more than 2 $\times$ faster but also achieves better generalization both for seen and unseen languages. Especially, it attains remarkable performance as unsupervised speech translation for Portuguese-English, achieving $8.2$ BLEU (compared to the supervised bilingual baseline $0.5$ BLEU as is provided in Table \ref{tab:m2o-romance}, and even beat ($+1.9$ BLEU) the previous state-of-the-art for this direction which is a supervised multilingual model. 


\noindent \textbf{Target-side (text) transfer.} Table \ref{tab:o2m-zero} show the proposed approach also achieves zero-shot transfer capability for translating \textit{to} a new languages, with unsupervised translation for English-Japanese only 1.3 BLEU gap compared to the best supervised result. Furthermore, an interesting finding is that applying LNA finetuning to decoder is crucial for zero-shot transfer to unseen languages (Ja), as finetuning the entire decoder tends to optimize the model on target languages seen during training. 


\begin{table*}
\begin{center}
\begin{tabular}{rc|cccccc}
\toprule
& & \multicolumn{6}{c}{\bf Target } \\
& & 
 De &  En &  Es & Fr   & It  &  Pt  \\ 
\midrule
\multirow{6}{*}{{\rotatebox[origin=c]{90}{\bf Source }}}
&  De & & \cg 12.8/\bf20.6 & 10.2/\bf13.8 & 11.6/\bf14.9 & 6.6/\bf8.6 & 10.4/\bf13.0 \\
&  En & \cg  13.1/\textbf{22.5*} & \cg  & \cg 23.1/\textbf{32.3*} & \cg 22.1/\textbf{30.0*} & \cg 14.9/\bf21.5 & \cg 20.7/\bf28.4 \\
&  Es & 9.2/\bf12.1 & \cg 18.9/\bf26.0 & & 19.0/\bf21.8 & 13.3/\bf15.4 &  20.0/\bf21.9 \\
&  Fr & 9.8/\bf13.6 & \cg 19.8/\textbf{27.9*} & 18.6/\bf21.7 & & 13.8/\bf15.2 & 19.7/\bf21.4 \\
&  It & 10.1/\bf11.9 & \cg 19.8/\bf25.6 & 18.8/\bf20.8 & 19.1/\textbf{20.0*} &  & 19.8/19.2 \\
&  Pt & 9.0/\bf11.4 & \cg 19.0/\bf24.1 & 19.8/19.6 & 18.1/\bf18.6 & 15.6/\bf16.1 &  \\
\bottomrule
\end{tabular}
\end{center}
\caption{Zero-shot performance (baseline/XMEF) on Europarl. Baseline is a many-to-many multilingual model trained on parallel data from all 30 directions. For our approach (XMEF), only to- and from- English directions (\colorbox{lightgray}{shaded}) were used in multilingual finetuning while the rest are results of zero-shot translation. \textbf{Bold} are where our model (En-only and zero-shot for the rest) outperforms a supervised many-to-many model. * means that our zero-shot model also beats the supervised cascade model in \cite{jairsan2020a}. }
\label{tab:zero-shot}
\end{table*}

\subsection{Multilingual Speech Translation}
We evaluate the performance of XMEF with multilingual finetuning on all 36 translation directions in CoVoST 2, respectively all 21 languages into English (many-to-one) and from English into 15 languages (one-to-many).

\textbf{Many to one.} Consistent with the observation of source-side crosslingual transfer in Sec \ref{subsec:fexl}, our approach (using self-supervised wav2vec 2.0 encoder pretraining on English only) perform very well on Romance, Germantic and Slavic language families in both high-resource ( $\ge$ 100 hours training data) and low-resource directions ($7\sim 44$ hours training data) as is summarized in Table \ref{tab:m2o-romance}, and even surpassing the best cascade results on 8 out of 10 languages. The resulting multilingual model also improves distant and extremely low resource (mostly $\le$ 5 hours training data) languages as is shown in Table \ref{tab:m2o-distant}. Especially, LNA-E,D outperforms (12.5 vs. 11.2 averaged BLEU) finetuning the entire model (Finetune All) with only training 20\% of the parameters. Applying LNA finetuning to decoder only (LNA-D) further improves the performance. Overall, all variants of XMEF can obtain one multilingual model which improved the previous state-of-the-art (SOTA) for 19 out of 21 directions.


\textbf{One to many.} Table \ref{tab:o2m-seen} summarizes performance on translating (from English) to 15 languages where multilingual models from XMEF have improved previous state-of-the-art (both E2E and cascade) on all directions  (+6.4 BLEU on average). The performance of applying LNA finetuning to encoder only (LNA-E) is very close to (24.2 vs. 24.5 averaged BLEU) that of finetuning the entire model (Finetune All) while has 40\% less parameters to train. Applying LNA to both encoder and decoder further reduces the amount of parameters to train to only $8\sim20$\% of all parameters in the pretrained models yet still maintain strong performance compared to strong baselines such as ASR PT with multilingual finetuning (ASR PT+Multi) as well as the best cascade models for 13 directions. The only two languages (Ca, Cy) did not get improved with LNA finetuning of decoder are due to that they were never seen during mBART self-supervised (monolinugual data only) pretraining nor ML501N pretraining with bitext.  

\textbf{Joint Training} In the many to one case,  language pairs with reasonable amount speech training data (+ 18 hours) and large amount of parallel text data (+1 million sentences) (``Fr-En", ``De-En", ``Es-En", ``It-En", ``Ru-En" and ``Fa-En"),  outperform the corresponding single task trained models and achieve the state-of-art results(~\autoref{tab:m2o-romance}). However, if the amount of speech data is too small (10 hours or less), joint training doesn't help too much and it even makes the performance worse(~\autoref{tab:m2o-distant}). In one to many case (``En-X"), where there are 364 hours English audio data for training, joint training improves the results further by another 0.6 BLEU~\autoref{tab:o2m-seen}.

\subsection{Zero-shot}


Finally, we evaluate how the proposed approach performs in zero-shot multilingual translation (translating X $\rightarrow$ Y after training on  X $\rightarrow$ En and En $\rightarrow$ Y. We apply LNA-D multilingual finetuning using En-X and X-En training data only from the Europarl corpus. Table \ref{tab:zero-shot} reports both the supervised performance on to- and from-English directions and zero-shot performance translating between non-Engligh languages without training on their parallel data. We compare to the strong baseline of a many-to-many multilingual model trained from scratch \textit{using all parallel data} from non-English directions as well as English-centric directions. Our approach improves both to- and from-English directions (+6.8 BLEU and +8.2 BLEU on averge respectively) and our zero-shot results also beats (+5.6 BLEU) the supervised many-to-many model on 28 pair-wise (except for It-Pt and Pt-Es) translation directions.



\section{Ablation Studies}
\label{sec:ablation}
\todo{Add a learning curve (train acc by time) to show fast adaptation.}



In Table \ref{tab:ablation-finetune} we analyze how components of LNA contribute to generalization performance and training efficiency. First, we found finetuning LayerNorm parameter (far less compared to the amount of multi-head attention parameters) is important for training stability when finetuning pretrained models without which (-LN) training diverges. 
Adding finetuning encoder attention (EA) parameters is important for adapting pretrained text decoder for ST task.
Finally, we found finetuning self attention (+SA) parameters in decoder did not bring further improvement while significantly increasing the amount of parameters to train.

\begin{table}[h]
\centering
\scalebox{0.9}{
\begin{tabular}{llccccccccccr}
\toprule
{\bf Enc } & {\bf Dec } & {\bf  PPL  $\downarrow$} & \textbf{Params (\%)}  \\
\midrule
LN  & LN + EA & 5.17 &   69.4M (8.8\%) \\
\midrule
- LN  &  - LN & 37.66 & 69.3M (8.7\%)\\
  &  - EA & 5.97 & 19.0M (2.4\%)\\
  &  + SA &  5.26 & 119.8M (15.1\%)\\
 + SA  &   & 5.53 & 170.2M (21.5\%)\\
\bottomrule
\end{tabular}
}
\caption{Ablation on LNA-Minimalist finetuning, where we evaluate the effect of finetuning LayerNorm (\textbf{LN}) and \textbf{A}ttention parameters. The experiment was conducted on the CoVoST English-German dataset and we report perplexity on the dev set. \textbf{\%} indicates the what percentage of total parameters of pretrained modules are trained during finetuning.}
\label{tab:ablation-finetune}
\end{table}

\section{Related Work}
\label{sec:related}

\subsection{Speech Translation}
Sequence-to-sequence based speech translation has shown very good potential over the traditional cascaded system~\cite{Berard2016ListenAT,Goldwater2017TowardsST, Weiss2017SequencetoSequenceMC} with end-to-end approaches surpassing cascaded system for the first time at IWSLT~\cite{ansari-etal-2020-findings}. However, previous work also indicates that its success heavily relies on large amounts of labelled training data, which is difficult to acquire. In order to mitigate the data scarcity issue, recent research work focuses on multi-task learning~\cite{Weiss2017SequencetoSequenceMC,Anastasopoulos2018TiedML,Bahar2019ACS,Wang2019BridgingTG,Wang2020CurriculumPF,Indurthi2020EndendST}, pre-training different components of the model~\cite{alex2018endtoend,bansal-etal-2019-pre}, transfer learning~\cite{gaido2020endtoend,Liu2019EndtoEndST} and generating synthetic data~\cite{Jia2018LeveragingWS,pino2020self}.
These methods aim to use weakly supervised data, i.e. speech-to-transcription or text-to-translation pairs in addition to fully supervised data, i.e. speech-to-translation pairs. 
 
\subsection{Pretraining}
This work is motivated by the recent success of self-supervised learning for NLP and speech processing applications~\cite{Radford2018ImprovingLU,bert,Clark2019ELECTRAPT,lewis2019bart,Lample2019CrosslingualLM,Dong2019UnifiedLM,Liu2020MultilingualDP,Tang2020MultilingualTW,Rivire2020UnsupervisedPT,Kawakami2020LearningRA,Chung2020ImprovedSR,Baevski2020wav2vecV2}. Masked language modeling~\cite{bert} is adopted to sequence to sequence framework~\cite{lewis2019bart,Dong2019UnifiedLM} and widely used as pre-training methods for multilingual translation task~\cite{Liu2020MultilingualDP,Tang2020MultilingualTW}. Contrastive Predictive coding~\cite{Oord2018RepresentationLW} is thriving for speech pre-training and has demonstrated its effectiveness on low resource ASR~\cite{Rivire2020UnsupervisedPT,Baevski2020wav2vecV2}. 
Initial results from \cite{wu2020selfsupervised} have shown the effectiveness of the speech self-supervised pre-training~\cite{Baevski2020vqwav2vecSL} for end-to-end ST. In this work, we aim to leverage pre-trained components from different modalities (text and speech) and applies them to the ST task. 
\subsection{Finetuning}
In NLP, finetuning large pretrained model is the state-of-the-art approach for various downstream tasks\cite{bert,raffel2020exploring}. However, given that most pretrained models contain large number of parameters. How to efficiently use them for downstream tasks have gained increased research interest. \cite{houlsby2019parameter} and \cite{pfeiffer2020mad} represents the stream of work which adds additional ``adaptor modules" to achieve fast adaptation to downstream tasks. Another category of solutions without adding parameters (which is where our work belongs to) is to finetune a subset of parameters for given downstream tasks. Empirical studies shows that finetuning the final layers of BERT account for most of the quality gains on downstream tasks \cite{kovaleva2019revealing,lee2019would}. Finetuning LayerNorm parameters was also found effective for adapting pretrained BART or mBART for (text-only) machine translation \cite{stickland2020recipes}. A general approach is to automatically learn which layers/parameters from a large-pretrained model to finetune and freeze \cite{guo2019spottune}, which we found is an exciting direction for future work.

\section{Conclusion}
\label{sec:majhead}

   
We proposed a simple and effective approach to leverage pretrained single-modality models (such as wav2vec 2.0, mBART, etc) to perform speech-to-text translation. 
On two large-scale multilingual speech translation benchmarks, our approach advances the state-of-the-art ($+5.9$ BLEU on average for 34 translation directions in CoVoST 2, and $+3.8$ BLEU for 28 translation directions in Europarl), achieved new state-of-the-art for end-to-end speech translation. We provide an efficient finetuning strategy which is not only data- and parameter-efficient, but also demonstrates crosslingual transfer ability by only finetuning $10\sim20$\% of the parameters of large pretrained models.

\bibliography{anthology,custom}
\bibliographystyle{acl_natbib}

\appendix
\section{Languages}

\begin{table}[h]
\begin{center}
\begin{tabular}[b]{lrcc}
\toprule
\textbf{Code} & 
\textbf{Language}  \\
\midrule
{\bf Ar }& Arabic  \\
{\bf Ca }& Catalan  \\
{\bf Cy }& Welsh \\
{\bf De}& German  \\
{\bf En }& English  \\
{\bf Et }& Estonian  \\
{\bf Es }& Spanish  \\
{\bf Fa }& Persian  \\
{\bf Fr }& French \\
{\bf Ja }& Japanese  \\
{\bf Id }& Indonesian \\
{\bf It }& Italian  \\
{\bf Lv }& Latvian  \\
{\bf Mn }& Mongolian  \\
{\bf Nl }& Dutch  \\
{\bf Pt }& Portuguese  \\
{\bf Ru }& Russian  \\
{\bf Sv }& Swedish  \\
{\bf Ta }& Tamil  \\
{\bf Tr }& Turkish  \\
{\bf Zh } & Chinese (Sim) \\
\bottomrule
\end{tabular}
\caption{ A list of 21 languages and their ISO codes with experiment results reported in this paper.}
\label{tab:datastats}
\end{center}
\end{table}

\section{Training Details}
\label{app:training}

When using wav2vec 2.0 encoder, we use 16-bit 16kHz mono-channel audios as inputs. When using a traditional speech recognition (ASR) encoder, we extract 80-channel log mel-filter bank features (25ms window size and 10ms shift) with utterance-level cepstral mean and variance normalization applied. We remove training samples with more than 3,000 frames for GPU memory efficiency.  We use the best checkpoint (without checkpoint averaging) according to validation loss and a beam size of 5 for decoding. We report case-sensitive detokenized BLEU using sacreBLEU~\cite{post-2018-call}, except for Japanese and Chinese translations (no word segmentation) where we report character-level BLEU. We implement all our experiments using fairseq S2T~\cite{ott2019fairseq,wang2020fairseqs2t}. 

 


\end{document}